\documentclass[runningheads]{llncs}

\usepackage{eccv}

\usepackage{eccvabbrv}

\usepackage{graphicx}
\usepackage{booktabs}
\usepackage{adjustbox}
\usepackage{subcaption}
\usepackage[figurename=Fig.]{caption}
\usepackage{tabularx} 
\usepackage[accsupp]{axessibility}  

\usepackage[pagebackref,breaklinks,colorlinks,citecolor=eccvblue]{hyperref}
\usepackage{hyperref}

\usepackage{orcidlink}

\begin{document}

\title{MobileUNETR: A Lightweight End-To-End Hybrid Vision Transformer For Efficient Medical Image Segmentation}

\titlerunning{MobileUNETR}

\author{Shehan Perera\inst{1}\orcidlink{0009-0005-3831-0404} \and
Yunus Erzurumlu\inst{1}\orcidlink{0009-0006-5798-5842} \and
Deepak Gulati\inst{2}\orcidlink{0000-0003-3374-5992} \and
Alper Yilmaz\inst{1}\orcidlink{0000-0003-0755-2628}}

\authorrunning{Perera et al.}

\institute{Photogrammetric Computer Vision Lab, The Ohio State University \\
Wexner Medical Center, The Ohio State University \\
\email{\{perera.27, yilmaz.15, erzurumlu.1\}@osu.edu}\\
\email{deepakkumar.gulati@osumc.edu}}

\maketitle

\begin{abstract}
Skin cancer segmentation poses a significant challenge in medical image analysis. Numerous existing solutions, predominantly CNN-based, face issues related to a lack of global contextual understanding. Alternatively, some approaches resort to large-scale Transformer models to bridge the global contextual gaps, but at the expense of model size and computational complexity. Finally many Transformer based approaches rely primarily on CNN based decoders overlooking the benefits of Transformer based decoding models. Recognizing these limitations, we address the need efficient lightweight solutions by introducing MobileUNETR, which aims to overcome the performance constraints associated with both CNNs and Transformers while minimizing model size, presenting a promising stride towards efficient image segmentation. MobileUNETR has 3 main features. 1) MobileUNETR comprises of a lightweight hybrid CNN-Transformer encoder to help balance local and global contextual feature extraction in an efficient manner; 2) A novel hybrid decoder that simultaneously utilizes low-level and global features at different resolutions within the decoding stage for accurate mask generation; 3) surpassing large and complex architectures, MobileUNETR achieves superior performance with 3 million parameters and a computational complexity of 1.3 GFLOP resulting in 10x and 23x reduction in parameters and FLOPS, respectively. Extensive experiments have been conducted to validate the effectiveness of our proposed method on four publicly available skin lesion segmentation datasets, including ISIC 2016, ISIC 2017, ISIC 2018, and PH2 datasets. The code will be publicly available at: \href{https://github.com/OSUPCVLab/MobileUNETR.git}{https://github.com/OSUPCVLab/MobileUNETR.git}.
\end{abstract}

\section{Introduction}
\label{sec1}

\begin{figure*}[ht]
  \centering
  \includegraphics[width=12cm, height=4cm]{"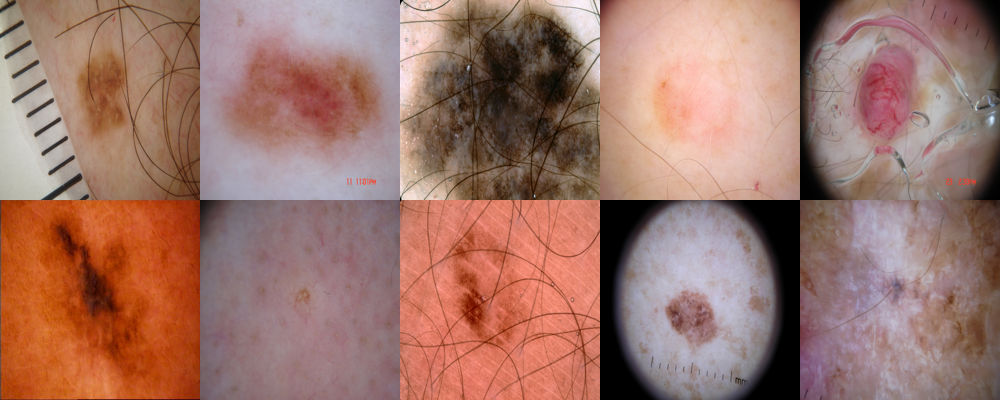"}
  \caption{Examples of typical skin cancer instances showcasing the typical noise and complicates in dermoscopic images.}
  \label{fig:examples}
\end{figure*}

Skin cancer, among the most prevalent and rapidly increasing forms of cancer worldwide, poses a significant global health challenge \cite{Siegel2017ColorectalCS}. Given the various forms of skin cancer that appear between patients and different levels of severity, accurately identifying and categorizing skin lesions becomes a complex task. One of the primary difficulties in diagnosing this form of cancer lies in visual inspection of the lesions. The subjective nature of the visual process, influenced by factors such as lighting conditions, individual expertise, and the inherent variability in the way skin cancer presents itself in different patients, make visual categorization a difficult task. To improve diagnostic precision, dermatologists use dermoscopy, a non-invasive technique for skin surface microscopy. Dermoscopy provides physicians with high-resolution images of the affected skin, allowing a closer examination of the characteristics of the lesion \cite{Haenssle2018ManAM}. Although this advancement has undoubtedly improved the accuracy of human visual analysis, it has not completely eliminated the challenges associated with human subjectivity. Dermatologists, even with the help of dermoscopic images, may still differ in their interpretation of skin lesions. This lack of consistency in diagnosis among medical professionals emphasizes the need for additional tools that can offer objective and standardized assessments. Recognizing these challenges, there has been a growing effort to integrate Computer-Aided Diagnostic (CAD) systems to support physicians in the diagnosis of skin lesions. 

Early iterations of CAD systems designed for skin cancer segmentation were often approached through complicated multi-step image processing pipelines \cite{Mishra2016AnOO}, \cite{EmreCelebi2008BorderDI, Tang2018AMF, Garnavi2011AutomaticSO}. Techniques employed in these early iterations include color-space transformations, principal component analysis, and the use of hand-crafted features, to name a few. Despite their progress in medical diagnosis, these approaches struggled to accurately delineate affected skin regions. Rule-based and hand-crafted systems often oversimplified complex, variable skin lesions, including artifacts and noise from body hair. 

The development of deep learning and its adoption represents a crucial step towards enhancing the efficiency and accuracy of CAD systems. These systems employ advanced neural networks to delineate the boundaries of the lesions, allowing a more precise assessment of their characteristics. Deep learning algorithms, with their ability to automatically learn intricate patterns and features directly from data, have demonstrated superior performance in segmenting skin lesion \cite{Wu2021FATNetFA, Goyal2020SkinLS, Xie2019AMB}]. These algorithms can discern subtle variations in color, texture, and shape, adapting dynamically to the diverse manifestations of skin cancer between different individuals.  

Central to the success of deep learning for medical image segmentation is the introduction of the encoder-decoder architecture. Encoder-Decoder architectures implemented via Fully Convolutional Neural Networks (FCNNs) have particularly excelled in this domain and have become the State-Of-The-Art (SOTA) for many segmentation tasks \cite{Yuan2017AutomaticSL, Ronneberger2015UNetCN}. Although highly successful, one of the major drawbacks of FCNN/CNN based approaches is their lack of long-range contextual understanding. Although CNNs excel at capturing local features within an image, they inherently struggle to gather broader context information or a global relationship between different elements. In particular, in the case of skin cancer, where lesions can vary significantly from patient to patient, a global understanding becomes crucial to help the model overcome ambiguities. To overcome context limitations within CNNs, researchers have resorted to larger and deeper models to help improve the overall receptive field through pure convolutions \cite{Schlemper2018AttentionGN, Lei2020SkinLS, Jha2019ResUNetAA}. However, this solution comes with its own set of challenges. Larger models require more computational resources, making them computationally expensive and slower to train and deploy. Additionally, the pursuit of a larger receptive field through sheer model size may lead to diminishing returns, emphasizing the need for a more efficient and effective approaches. The integration of self-attention modules introduced in the Transformer \cite{Vaswani2017AttentionIA} architecture with convolutional layers has been suggested as a means to enhance the non-local modeling capability \cite{Hatamizadeh2022SwinUS, perera2024segformer3d} and offer promising long-range contextual understanding benefits for many downstream tasks. 

Originally developed for Natural Language Processing (NLP), the Transformer architecture has seen significant adoption to many computer vision tasks. With the initial Vision Transformer \cite{Dosovitskiy2020AnII} that allowed Transformers to perform image classification, researchers were provided with an architecture that is capable of modeling long-range dependencies and gathering global context clues at every stage of the model. However, as a trade-off the self-attention mechanism, central to the Transformer architecture, proves computationally expensive, especially at large spatial dimensions. Additionally, ViTs produce single-scale features, in contrast to multi-scale features typically generated by CNN models \cite{Xiao2021EarlyCH}. This trade-off between global awareness and computational efficiency presents a significant challenge when employing transformer architectures in resource-limited real-world applications.

\begin{figure*}
  \centering
  \includegraphics[width=\textwidth, height=6cm]{"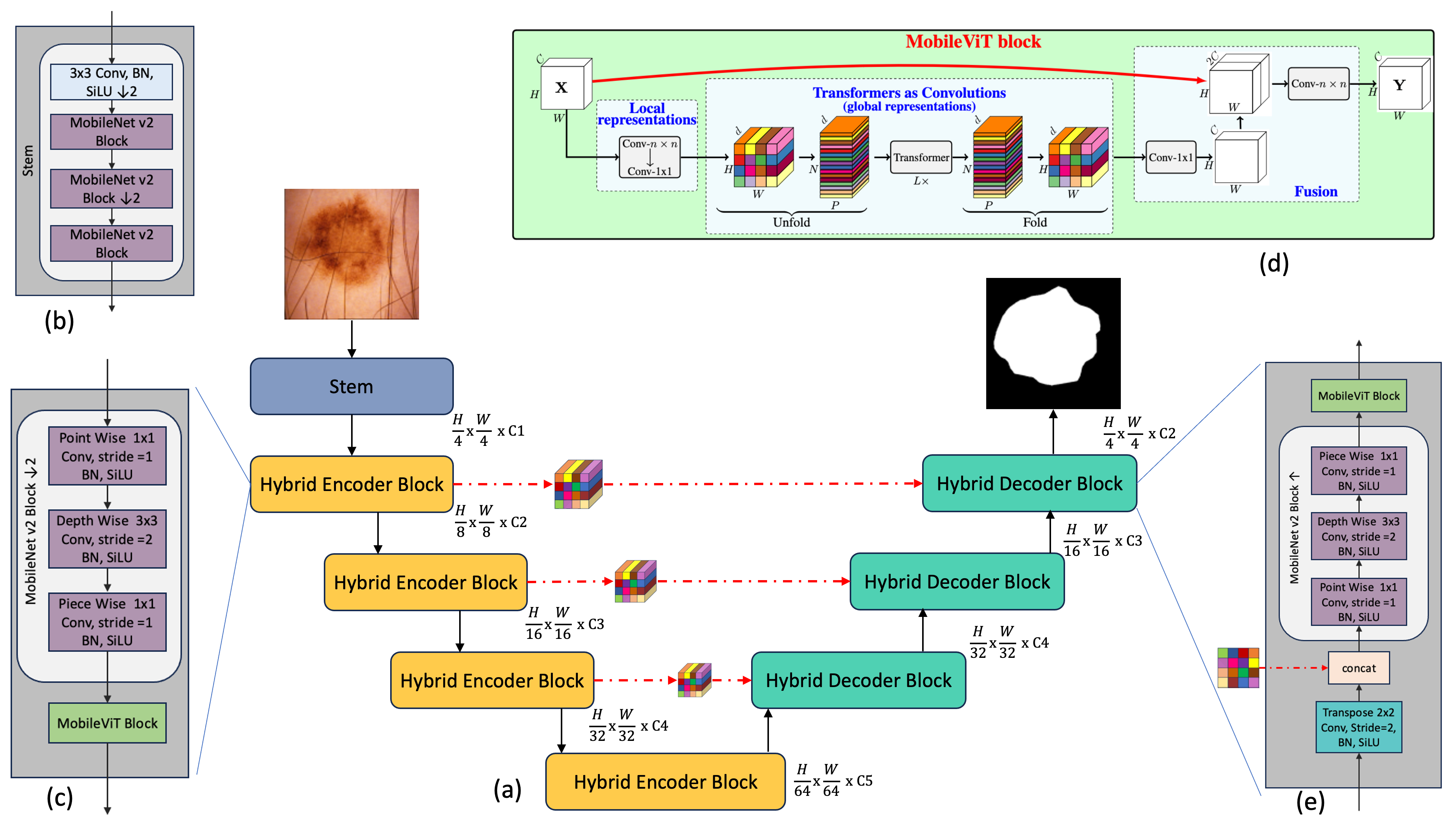"}
  \caption{MobileUNETR Architecture: (a) Main MobileUNETR architecture showcases a hierarchical hybrid encoder-decoder architecture to extract and combine coarse and fine-grained features in an end-to-end framework. (b) Lightweight convolution stem for low-level feature extraction and spatial downsampling. (c) Hybrid Encoder Block efficiently extracts local and global features at each stage. (d) MobileViT Block for global feature extraction and long-range context understanding. (e) Novel Decoder Block for efficient upscaling and combining local/global features while allowing the model to dynamically adapt features during the decoder stage.}
  \label{fig:architecture}
\end{figure*}

To overcome current bottlenecks in widely adopted CNN and Transformer architectures we introduce MobileUNETR, a novel end-to-end transformer based encoder decoder architecture for efficient image segmentation. At a high level, challenging and complex image segmentation tasks often benefit from feature extraction capabilities that consider local and global contextual information within the feature encoding stage. However, segmentation approaches typically focus on optimizing the feature extractor while overlooking the importance of developing novel decoding strategies. Common segmentation frameworks in medical imaging, utilizing complex CNN and/or Transformer structures, generally favor excluding Transformer based decoders, opting instead for pure CNNs \cite{Hatamizadeh2021UNETRTF, Hatamizadeh2022SwinUS, Shaker2022UNETRDI, Lee20223DUA}. This choice can be attributed to the fact that, despite being great at capturing global information, Transformers are unable to capture intricate local details which are highly useful when generating accurate segmentation masks. To overcome the over-reliance on pure CNN layers within the decoder stage, we propose a novel highly effective and light-weight decoder capable of learning and integrating local/global details to generate highly accurate segmentation masks. 

We demonstrate the advantages of MobileUNETR in terms of model size, run time complexity, and accuracy on four publicly available skin lesion segmentation datasets, including ISIC 2016 \cite{Gutman2016SkinLA}, ISIC 2017 \cite{Gutman2017SkinLA}, ISIC 2018 \cite{Codella2019SkinLA}, and PH2 \cite{Mendona2013PH2A} datasets. We demonstrate a significant increase in performance across all datasets and advanced architectures and training methodologies while reducing the model size and complexity by 10x and 23x, respectively.

Our main contributions can be summarized as follows. 

\begin{enumerate}
\item We propose a novel lightweight and efficient end-to-end Transformer based hybrid model for skin lesion segmentation, where local and global contextual features are enforced at each stage to retain global awareness of a given scene.   

\item To overcome the over-reliance on CNN based decoding strategies we introduce a novel Transformer based hybrid decoder that simultaneously utilize low-level and global features at different resolutions for highly accurate and well aligned mask generation.

\item The proposed architecture surpasses large and highly complex CNN, Transformer, and Hybrid models in segmentation with only 3 millions parameters and 1.3 GFLOP of complexity resulting in a 10x and 23x reduction in computational complexity, respectively, than the current SOTA models. 

\end{enumerate}

\section{Related Works}

Skin lesion segmentation is critical in automated dermatological diagnosis; however, it is difficult due to lesion diversity and the presence of noise in the images. Traditional image processing methods have given way to advanced deep learning systems, particularly Convolutional Neural Networks (CNNs), and then Transformer-based methods, which have considerably improved segmentation accuracy and reliability.

\subsection{CNN Based Methods}

After the increasing popularity of Deep Neural Networks (DNNs) and Convolutional Neural Networks (CNNs), they have become the go-to tools for skin lesion segmentation tasks. They have creatively solved difficulties such as feature discernment and data variability management. The field has seen notable developments, such as the introduction of a multistage fully convolutional network (FCN) to the field by \cite{Bi2017DermoscopicIS}, which incorporates a parallel integration method to enhance the segmentation of skin lesions' boundaries. \cite{Yuan2017ImprovingDI} have contributed similarly by creating an improved convolutional-deconvolutional network specifically optimized for dermoscopic image analysis and integrating various color spaces to better diagnose lesions. \cite{Jha2020DoubleUNetAD} expanded on this trend of architectural innovation with their DoubleU-Net, which combines multiple U-Net structures to improve segmentation accuracy. 

In parallel, efforts to develop automated detection systems have been prominent. \cite{Ratul2019SkinLC} worked on the early detection of malignant skin lesions using dilated convolutions across multiple architectures such as VGG16, VGG19 by \cite{Simonyan2014VeryDC}, MobileNet by \cite{Howard2017MobileNetsEC}, and InceptionV3 by \cite{Szegedy2015RethinkingTI}, as well as the HAM10000 dataset by \cite{Tschandl2018TheHD}  for training and testing. The use of pre-trained networks and deep learning models is also evident in the multiple winning solutions at the ISIC 2018 Challenge \cite{Adegun2020DeepLT}, where many built their model on the DeepLab \cite{Chen2016DeepLabSI} architecture using pre-trained weights from PASCAL VOC-2012 \cite{Everingham2014ThePV} and used ensemble approach's among others with models such as VGG16, U-net, DenseNet by \cite{Huang2016DenselyCC}, and Inceptionv3, fine-tuning these with additional training iterations for state-of-art performance. 

\cite{Taghanaki2018SelectAA} and \cite{Tang2018AMF} proved the flexibility of these models in varied context-aware settings by improving feature extraction in CNNs, the former through modified skip connections and the latter with multistage UNets. Furthermore, \cite{Abraham2019ANF} introduced a new focal Tversky loss function to address data imbalance, a significant difficulty in medical imaging, improving the precision recall balance for small lesion structures.

The ISIC 2019 challenge also led to several new studies using CNNs for dermoscopic medical imaging. \cite{Pollastri2019AugmentingDW}, \cite{Pacheco2019SkinCD}, and \cite{To2019EnsembledSC} used a variety of CNN architectures with different data augmentation methods. These studies demonstrated CNNs ability to segment skin lesions locally, but their performance shortcomings can be attributed to their inability to extract valuable global context information.

\subsection{Transformer Based Methods}

Being limited to only local features forced researchers to seek new approaches. This caused an evolution towards the usage of global feature-based tools. This evolution is distinguished by a shift from standard CNN-based techniques toward novel ways of using transformers and self-attention mechanisms. \cite{Li2019DenseDN} pioneered the Dense Deconvolutional Network (DDN) in skin lesion segmentation, employing dense layers and chained residual pooling to capture long-range relationships, a significant departure from prior approaches. Furthermore, \cite{Xue2018AdversarialLW} investigated adversarial learning with SegAN, improving segmentation accuracy by adeptly capturing subtle relationships, a significant development in dermatological imaging. \cite{Mirikharaji2018StarSP} and \cite{Feng2020CPFNetCP} substantially advanced skin lesion segmentation with their new methodologies. Mirikharaji and Hamarneh implemented a star-shape prior (SSP) in a fully convolutional network to improve accuracy and reliability by penalizing non-star-shaped regions while preserving global structures. The use of shape priors to segment complex skin lesion patterns was demonstrated in this study. \cite{Feng2020CPFNetCP} supplemented this with CPFNet, which uses pyramidal modules to collect global context in feature maps, successfully managing skin lesion variability and enhancing delineation accuracy in intricate lesion patterns. 

Using transformers in neural networks, pioneered by \cite{Vaswani2017AttentionIA}, was a significant turning point. \cite{Carion2020EndtoEndOD} and \cite{Dosovitskiy2020AnII} introduced transformers to computer vision. \cite{Zhao2020ExploringSF} demonstrated the effectiveness of self-attention mechanisms in image recognition models, which is helpful for complex skin lesion patterns. Additionally, \cite{Chen2021TransUNetTM} created TransUNet, which combines Transformers and U-Net to improve medical picture segmentation. The strength of TransUNet is in effectively encoding picture patches from CNN feature maps, which is essential for capturing detailed global context in segmentation tasks. \cite{Gulzar2022SkinLS} demonstrated TransUNet's performance in skin lesion segmentation, stressing its superior accuracy and dice coefficient over standard models, emphasizing the benefits of merging CNNs with transformers in medical imaging. Moreover, \cite{Wang2021BoundaryAwareTF} developed the boundary-aware Transformer (BAT) for segmentation of skin lesions. 

\begin{figure*}[ht]
  \centering
  \includegraphics[width=10cm, height=5cm]{"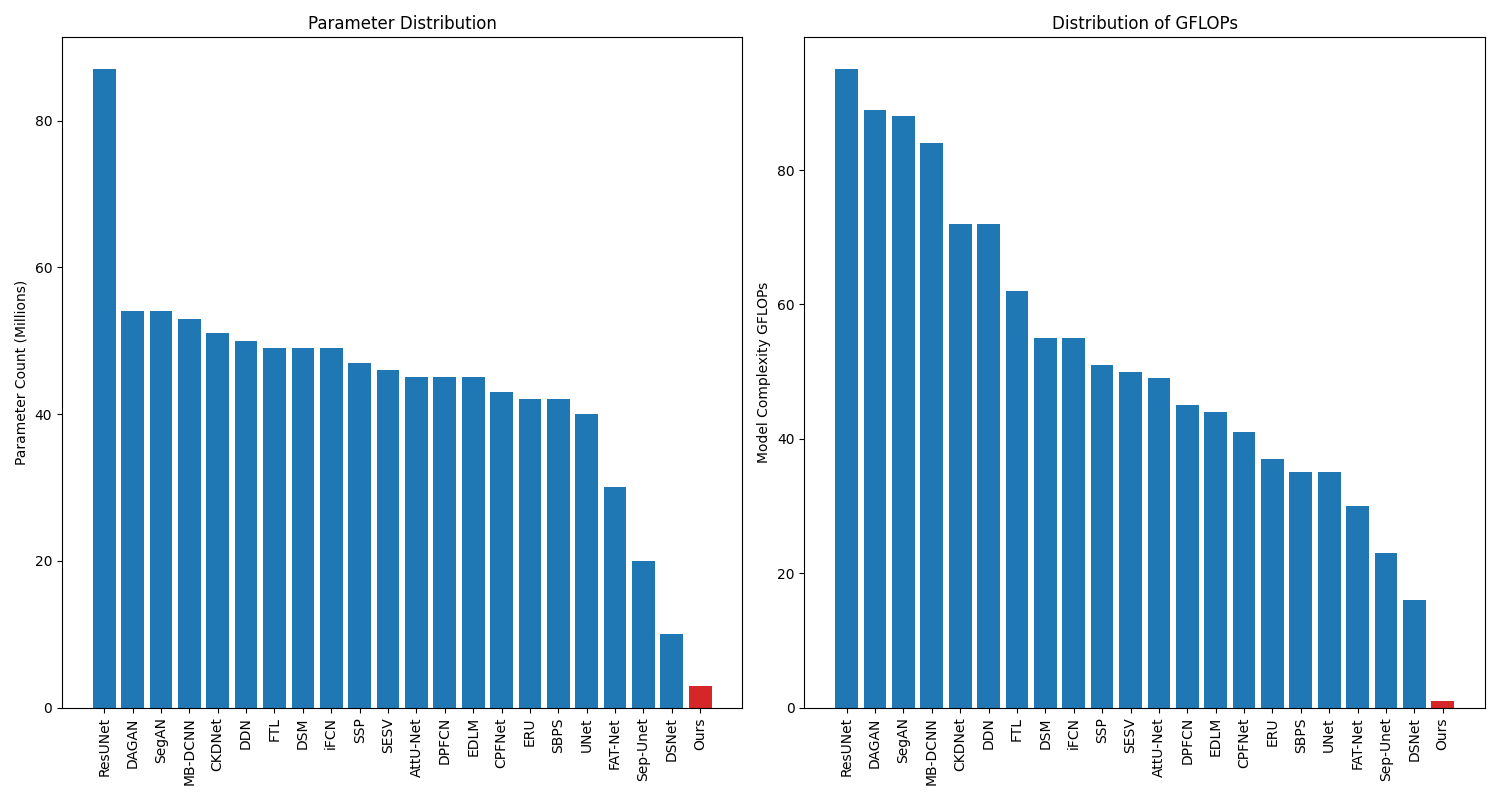"}
  \caption{Parameter count and GFLOP distributions (smaller is better) spanning SOTA models ranging from CNN, Transformer and Hybrid Architectures. We demonstrate a significant reduction in both model size and computational complexity compared to current SOTA architectures while achieving superior performance.}
  \label{fig:flops}
\end{figure*}

BAT incorporates a boundary-wise attention gate in its transformer structure to address unclear lesion boundaries, efficiently collecting global and local information in skin lesion imaging. FAT-Net, a feature-adaptive transformer network for segmentation of skin lesion, was introduced by \cite{Wu2021FATNetFA}. FAT-Net adeptly maintains long-range dependencies and contextual nuances by incorporating an extra transformer branch into the standard encoder-decoder structure, precisely addressing the variability and irregularity in skin lesions and improving melanoma analysis.

\section{Methodology}

In this section, we introduce MobileUNETR, our high-performance, efficient, and lightweight architecture for skin lesion segmentation. As shown in Figure \ref{fig:architecture}, the core MobileUNETR architecture consists of two main modules: (1) First, a lightweight hybrid encoder that efficiently generates coarse high-level and fine-grained low-level features; and (2) A novel lightweight hybrid decoder that effectively combines multilevel features while factoring in local and global context clues to generate high-accuracy semantic segmentation masks.

\subsection{Model Complexity}

The overarching goal of the medical imaging community is the pursuit of performance over complexity on a particular task such as skin lesion segmentation. One of the main contributions of the proposed MobileUNETR architecture is to demonstrate that well-constructed lightweight and efficient models can offer much better performance compared to large computationally expensive architectures. As seen in Figure \ref{fig:flops}, the proposed architecture is 10X smaller and 23X more computationally efficient against SOTA architectures in skin lesion segmentation while generating better results. Simplifying the model not only enhances training and performance on small datasets but also facilitates deployment in resource-limited environments.

\subsection{Encoder}

Two major groups of deep learning architectures exist in medical vision research, CNNs and Transformers, each with their own advantages and disadvantages. CNNs have been the defacto approach for many medical vision applications due to its efficiency, natural inductive biases and its ability to hierarchically encode features. However, despite its success, pure CNN based feature encoders are unable to effectively gain global contextual understanding of a given scene. Many hand-crafted approaches have been proposed to help CNNs obtain a larger receptive field such as dilated convolutions \cite{Chen2017RethinkingAC} and deeper models, however, image size and computational complexity constraints further research is required to help improve overall performance. Unlike CNNs, Transformers are designed to achieve a true global understanding of a scene. However, their computational constraints at large spatial resolutions hinder their adoption for efficient deep learning applications. 

By exploiting the natural advantages and disadvantages of CNNs and Transformer architectures, our proposed encoder maximizes feature representation capabilities while significantly minimizing computational complexity and parameter count.  At a high level, the feature extraction modules can be broken down into two stages: 1) CNN based local feature extraction and downsampling, 2) Hybrid Transformer/CNN based local and global representation learning.

{\bf \em CNN based local feature extraction}: End-to-end transformer models for computer vision, such as ViT and its derivatives \cite{Kolesnikov2019BigT, Zhou2021nnFormerIT, Liu2021SwinTH} result in large computationally complex models due to large sequence lengths generated for each input image. By combining the sequence length bottleneck in Transformers and the natural tendency of ViTs to learn low-level features in early layers \cite{Raghu2021DoVT}, simple CNN based early feature extraction substitution can be added to significantly reduce the computational complexity of the architecture. Specifically, MobileNet \cite{Howard2017MobileNetsEC} downsampling blocks are used within the proposed architecture to minimize the computational complexity of the low-level feature extraction stage without compromising the learned feature representations. Additionally, CNN based features allow the model to better incorporate spatial information compared to pure ViT based approaches while effectively reducing the spatial dimensions of the input data, allowing downstream transformer layers to efficiently learn global feature representations. 


{\bf \em Hybrid Transformer/CNN blocks}: Once efficient down-sampling is performed via CNNs to mitigate the computational complexity associated with large spatial resolutions, the MobileViT block is used to simultaneously extract local and global representations. The MobileViT block allows us to incorporate the long-range contextual benefits of Transformers while maintaining spatial ordering and local inductive biases. The operation can be broken down into two main components, as seen in Figure \ref{fig:architecture}. First, CNN-based depth-wise separable convolution \cite{Howard2017MobileNetsEC} is applied to encode spatial information and project features into high-dimensional space. Finally, to model long-range dependencies, the tensor is unfolded into non-overlapping flattened patches, and self attention layers are applied to capture interpatch relationships. This combination allows each feature map to have local and global understanding of the scene at each stage, improving its contextual understanding of the scene. 

\subsection{Decoder}

Most segmentation models emphasize the importance of the encoder stage for great segmentation performance. Here, local and global understanding is favored to ensure that relevant features that contain both are learned, compressed, and passed forward to the next stage. Most encoder-decoder approaches that use CNNs, Transformers, or CNN/Transformer dual encoders for feature extraction heavily rely on pure convolution to map extracted features to the final segmentation mask. A drawback of this approach is that, by using pure CNN layers within the decoder, we force the model to use information extracted at the bottleneck to learn features that ensure local continuity without providing it the capability to recalibrate itself using global contextual information. Additionally, naively stacking CNN layers can lead to large decoder modules, adding to the overall computational complexity of the encoder-decoder architecture. Our novel hybrid decoder architecture is a fast, computationally efficient, and lightweight approach that allows the model to hierarchically construct the final segmentation mask while ensuring that local and global context features are used at every stage of the decoding process. The proposed decoder module at ~1.5 million parameters efficiently combine the benefits of CNN and Transformer architectures, into an alternative to CNN based decoding methods. 

{\bf \em Simple Hybrid Decoder}: Typical CNN decoder modules in medical imaging \cite{Hatamizadeh2021UNETRTF, Wu2021FATNetFA, Hatamizadeh2022SwinUS} extract and refine the features of the encoder with a combination of transpose and standard convolutions. Using this structure allows the model to hierarchically increase the spatial resolution while refining features at each level with the help of features provided via skip connections. Despite their success, decoders that rely solely on CNNs face challenges in dynamically adapting their own features to ensure that the features learned at each stage are globally aligned. The proposed lightweight decoder performs three operations at each stage to ensure that the features extracted at each decoding stage are locally and globally aligned Figure \ref{fig:architecture}e. First, the feature map from the previous stage is upsampled via transpose convolutions. Next, we perform a local refinement of the upsampled features by combining information with the respective skip connections. Finally, the Transformer/CNN hybrid layers are used to allow the model to dynamically adjust itself based on long-range global contexts. By combining local refinement with global refinement stages, we allow the decoder to generate features that improve segmentation results by improving both local and global boundaries that are well aligned at each stage. 

\begin{figure*}[ht!]
    \centering
    \begin{subfigure}{10cm}
    \centering
    \captionsetup{labelformat=empty} 
    \caption{\textbf{Table 1:} Results for ISIC 2016 Dataset. MobileUNETR showcases significant advantages on parameter count, flops, and segmentation performance}
    \begin{adjustbox}{width=10cm, center}
    \setlength{\tabcolsep}{5pt} 
    \begin{tabular}{llccccccccc}
    \toprule
    {} &                                 Method &    SE &     SP &    ACC &    IoU &   Dice &  Params &  GFLOPs \\
        \midrule
    &      UNet \cite{Ronneberger2015UNetCN} &  90.16 &  96.56 &  94.66 &  81.84 &  88.84 &    40.0 &    89.0  \\
    &                DDN \cite{Li2019DenseDN}  &  92.61 &  96.25 &  95.05 &  84.43 &  90.52 &    50.0 &    49.0  \\
    &    AttU-Net \cite{Schlemper2019AttentionGN}  &  90.31 &  96.45 &  94.14 &  81.58 &  88.75 &    45.0 &    84.0  \\
    &   DPFCN \cite{NasrEsfahani2019DensePL}  &  91.50 &  96.12 &  94.93 &  84.12 &  89.24 &    45.0 &    88.0  \\
    &  Separable-Unet \cite{Tang2019EfficientSL}  &  \textbf{93.14} &  94.68 &  95.67 &  84.27 &  89.95 &    20.0 &    37.0  \\
    &              SBPS \cite{Lee2020StructureBP}  &  92.43 &  96.13 &  94.96 &  84.34 &  90.42 &    42.0 &    41.0  \\
    &           CPFNet \cite{Feng2020CPFNetCP}  &  92.11 &  95.91 &  95.09 &  83.81 &  90.23 &    43.0 &    16.0  \\
    &             DAGAN \cite{Lei2020SkinLS}  &  92.28 &  95.68 &  95.82 &  84.42 &  90.85 &    56.0 &    62.0  \\
    &               FAT-Net \cite{Wu2021FATNetFA}  &  92.59 &  96.02 &  96.04 &  85.30 &  91.59 &    30.0 &    23.0  \\
    \midrule
    &  \textbf{Ours}  &  93.03 &  \textbf{96.87} &  \textbf{96.59} &  \textbf{87.47} &  \textbf{92.80} &     \textbf{3.0} &     \textbf{1.3} \\
    \bottomrule
  \end{tabular}
    \label{tab: table1}
    \end{adjustbox}
    \end{subfigure}
    \bigskip
    \begin{subfigure}{10cm}
    \centering
    \includegraphics[width=8cm, height=4cm]{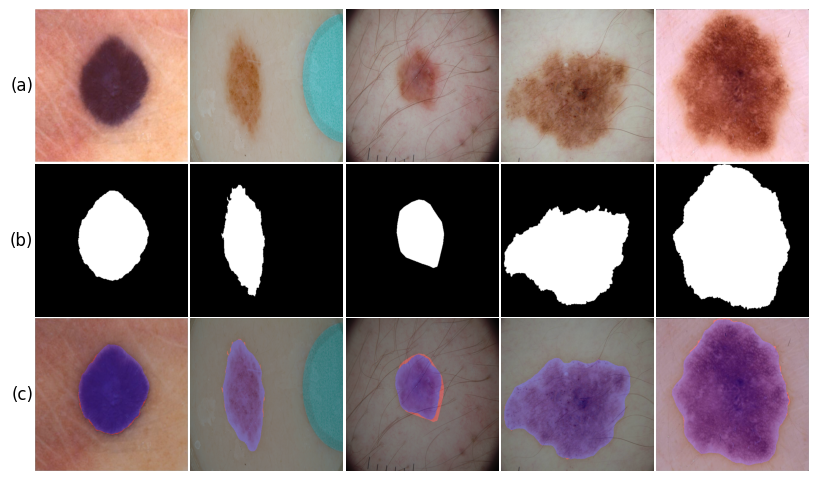}
    \captionsetup{labelformat=empty} 
    \caption{\textbf{Fig. 4:} Qualitative results on ISIC 2016 Dataset. (a) Original dermoscopic input (b) Ground truth mask (c) Overlaid predicted mask (blue) and ground truth mask (red) on original image. Qualitative results complements the quantitative findings in terms of segmentation performance. Appearance of red in the overlaid image indicates segmentation differences between the predicted and ground truth.}
    \label{fig: table1fig}
    \end{subfigure}
\end{figure*}

\vspace{-15pt}
\section{Experimental Results}

To showcase MobileUNETR's effectiveness as a highly competitive segmentation architecture, we perform multiple experiments across widely popular skin lesion segmentation datasets as well as compare and contrast its performance of the proposed model against high performing segmentation models. 

\vspace{-5pt}
\subsection{Dataset}

\begin{figure*}[ht]
    \centering
    \begin{subfigure}{10cm}
        \centering
            \captionsetup{labelformat=empty} 
        \caption{\textbf{Table 2:} Results for ISIC 2017 Dataset. MobileUNETR showcases significant advantages on parameter count, flops, and segmentation performance}
    \begin{adjustbox}{width=10cm, center}
    \setlength{\tabcolsep}{5pt} 
    \begin{tabular}{llccccccccc}
    \toprule
    {}                                   Method &   SE &     SP &    ACC &    IoU &   Dice &  Params &  GFLOPs \\
    \midrule
           UNet \cite{Ronneberger2015UNetCN}  &  81.72 &  96.80 &  91.64 &  72.34 &  81.59 &    40.0 &    89.0  \\
      SSP \cite{Mirikharaji2018StarSP}  &  83.18 &  97.15 &  92.14 &  75.21 &  83.04 &    47.0 &    72.0  \\
                  SegAN \cite{Xue2018AdversarialLW}  &  83.42 &  95.92 &  92.86 &  75.56 &  83.95 &    54.0 &    50.0  \\
                    DDN \cite{Li2019DenseDN}  &  83.64 &  95.97 &  92.35 &  75.27 &  84.29 &    50.0 &    49.0   \\
         AttU-Net \cite{Schlemper2019AttentionGN}  &  79.98 &  \textbf{97.76} &  91.45 &  71.73 &  80.82 &    45.0 &    84.0  \\
                  DSM \cite{Zhang2019DSMAD}  &  83.72 &  96.58 &  92.86 &  75.72 &  84.15 &    49.0 &    45.0  \\
                CPFNet \cite{Feng2020CPFNetCP}  &  83.44 &  96.45 &  92.15 &  75.46 &  84.03 &    43.0 &    16.0  \\
                ERU \cite{Nguyen2020SkinLS}  &  82.97 &  96.62 &  91.98 &  75.18 &  84.13 &    42.0 &    35.0   \\
                  SESV \cite{Xie2020SESVAM}  &  83.26 &  96.68 &  92.23 &  75.31 &  83.92 &    46.0 &    30.0  \\
              MB-DCNN \cite{Xie2019AMB}  &  83.25 &  96.84 &  93.11 &  76.03 &  84.27 &    53.0 &    55.0   \\
                 DAGAN \cite{Lei2020SkinLS}  &  83.63 &  97.16 &  93.04 &  75.94 &  84.25 &    56.0 &    62.0   \\
                   FAT-Net \cite{Wu2021FATNetFA}  &  83.92 &  97.25 &  93.26 &  76.53 &  85.00 &    30.0 &    23.0   \\
                   \midrule
         \textbf{Ours} &  \textbf{85.18} &  96.93 &  \textbf{94.46} &  \textbf{79.00} &  \textbf{86.84} &     \textbf{3.0} &     \textbf{1.3}  \\
    \bottomrule
  \end{tabular}
    \label{tab: table2}
    \end{adjustbox}

    \end{subfigure}
    \bigskip
    \begin{subfigure}{10cm}
    \centering
    \includegraphics[width=8cm, height=4cm]{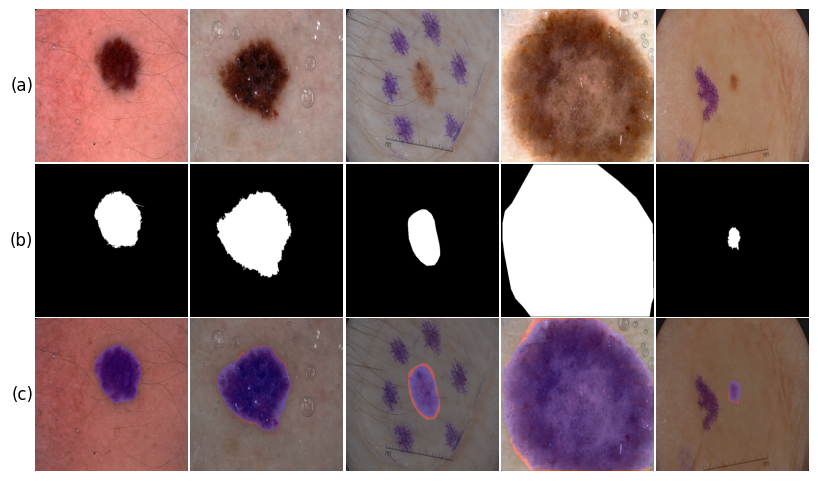}
    \captionsetup{labelformat=empty} 
    \caption{\textbf{Fig. 5:} Qualitative results on the ISIC 2017 Dataset. (a) Original dermoscopic input (b) Ground truth mask (c) Overlaid predicted mask (blue) and ground truth mask (red) on original image. Qualitative results complements the quantitative findings in terms of segmentation performance.Appearance of red in the overlaid image indicates segmentation differences between the predicted and ground truth.}
    \label{fig: table2fig}
    \end{subfigure}    
\end{figure*}

\begin{figure*}[ht]
    \centering
    \begin{subfigure}{10cm}
    \centering
    \captionsetup{labelformat=empty} 
    \caption{\textbf{Table 3:} Results for the ISIC 2018 Dataset. MobileUNETR showcases significant advantages on parameter count, flops, and segmentation performance}
    \begin{adjustbox}{width=10cm, center}
    \setlength{\tabcolsep}{5pt} 
    \begin{tabular}{llccccccccc}
    \toprule
    {} &                               Method &      SE &     SP &    ACC &    IoU &   Dice &  Params &  GFLOPs \\
    \midrule
    &    UNet \cite{Ronneberger2015UNetCN}  &  88.00 &  96.97 &  94.04 &  77.33 &  85.45 &    40.0 &    89.0 \\
    &  AttU-Net \cite{Schlemper2019AttentionGN}  &  86.00 &  \textbf{98.26} &  93.76 &  77.64 &  85.66 &    45.0 &    84.0 \\
    &       ResUNet ++ \cite{Jha2019ResUNetAA}  &  87.35 &  97.21 &  93.82 &  77.21 &  85.36 &    87.0 &    95.0 \\
    &       FTL \cite{Abraham2019ANF} &  87.54 &  96.32 &  94.12 &  78.25 &  86.93 &    49.0 &    55.0 \\
    &         CPFNet \cite{Feng2020CPFNetCP}  &  89.53 &  96.55 &  94.96 &  79.88 &  87.69 &    43.0 &    16.0 \\
    &          ERU \cite{Nguyen2020SkinLS}  &  90.32 &  96.92 &  94.35 &  80.56 &  88.12 &    42.0 &    35.0 \\
    &           DAGAN \cite{Lei2020SkinLS}  &  90.72 &  95.88 &  93.24 &  81.13 &  88.07 &    54.0 &    62.0 \\
    &          CKDNet \cite{Qiangguo2021CascadeKD}  &  90.55 &  97.01 &  94.92 &  80.41 &  87.79 &    51.0 &    44.0 \\
    &             FAT-Net \cite{Wu2021FATNetFA}  &  91.00 &  96.99 &  \textbf{95.78} &  82.02 &  89.03 &    30.0 &    23.0 \\
    \midrule
    &  \textbf{Ours} &  \textbf{92.55} &  95.03 &  94.40 &  \textbf{84.56} &  \textbf{90.74} &     \textbf{3.0} &     \textbf{1.3} \\
    \bottomrule
  \end{tabular}
    \label{tab: table3}
    \end{adjustbox}
    \end{subfigure}
    \bigskip
    \begin{subfigure}{10cm}
    \centering
    \includegraphics[width=8cm, height=4cm]{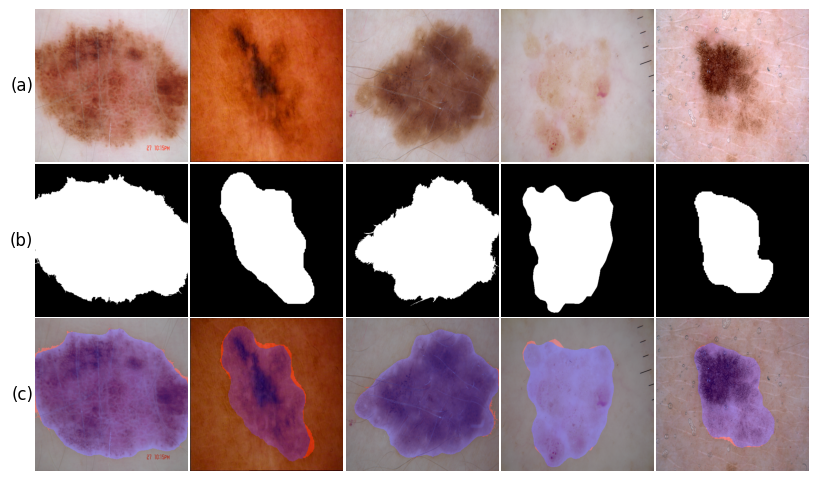}
    \captionsetup{labelformat=empty} 
    \caption{\textbf{Fig. 6:} Qualitative results for the ISIC 2018 Dataset. (a) Original dermoscopic input (b) Ground truth mask (c) Overlaid predicted mask (blue) and ground truth mask (red) on original image. Qualitative results complements the quantitative findings in terms of segmentation performance.Appearance of red in the overlaid image indicates segmentation differences between the predicted and ground truth.}
    \label{fig: table3fig}
    \end{subfigure}
\end{figure*}

To evaluate the performance of our efficient and lightweight model, MobileUNETR, we used four publicly available datasets for segmenting skin lesions. The International Skin Imaging Collaboration (ISIC) has developed and released three widely used datasets ISIC 2016, ISIC 2017 and ISIC 2018 for the task of skin lesion segmentation. Additionally, we evaluated our model performance on the PH2 dataset made available by Dermatology Service of Hospital Pedro Hispano, Portugal. The dataset breakdowns are provided below. 

\subsection{Implementation Details}

Our proposed MobileUNETR and accompanying experiments are trained and evaluated using PyTorch on a server equipped with a CPU and RTX 3090 GPU. All models follow a simple training procedue with AdamW \cite{Kingma2014AdamAM} optimizer with parameters $B1 = 0.9$ and $B2 = 0.999$, employing a batch size of 8. The experimental setup incorporates a linear warming-up stage spanning 40 epochs, during which the learning rate gradually increases from 0.0004/40 to 0.0004. Subsequently, a cosine annealing scheduler is employed to decay the learning rate over 400 epochs. Adhering to established practices, we employ straightforward data preparation and augmentation techniques available in PyTorch, ensuring the accessibility and reproducibility of our results.
\vspace{-10pt}
\subsection{Results on ISIC 2016}

\begin{figure*}[ht]
    \centering
    \begin{subfigure}{10cm}
    \centering
    \captionsetup{labelformat=empty} 
        \caption{\textbf{Table 4:} Results for the PH2 Dataset. MobileUNETR presents significant advantages on parameter count, flops, and segmentation performance}
    \begin{adjustbox}{width=10cm, center}
    \setlength{\tabcolsep}{5pt} 
    \begin{tabular}{llccccccccc}
    \toprule
    {} &                                 Method &     SE &     SP &    ACC &    IoU &   Dice &  Params &  GFLOPs \\
    \midrule
    &      UNet \cite{Ronneberger2015UNetCN}  &  91.25 &  95.88 &  92.33 &  84.10 &  89.36 &    40.0 &    89.0 \\
    &    AttU-Net \cite{Schlemper2018AttentionGN}  &  92.05 &  96.40 &  92.76 &  85.82 &  90.03 &    45.0 &    84.0 \\
    &             DSM \cite{Zhang2019DSMAD}  &  89.95 &  96.33 &  93.12 &  88.96 &  92.31 &    49.0 &    45.0 \\
    &            EDLM \cite{Goyal2020SkinLS}  &  92.36 &  94.83 &  94.52 &  85.34 &  91.81 &    45.0 &    51.0 \\
    &  Separable-Unet \cite{Tang2019EfficientSL}  &  \textbf{96.33} &  95.64 &  95.92 &  88.81 &  93.02 &    20.0 &    37.0 \\
    &           DSNet \cite{Hasan2019DSNetAD}  &  96.01 &  96.08 &  94.82 &  87.15 &  91.97 &    10.0 &    35.0 \\
    &       iFCN \cite{ztrk2020SkinLS}  &  96.13 &  95.91 &  96.08 &  87.56 &  93.21 &    49.0 &    72.0 \\
    &          MB-DCNN \cite{Xie2019AMB}  &  95.35 &  95.26 &  95.87 &  87.12 &  93.25 &    53.0 &    55.0 \\
    &               FAT-Net \cite{Wu2021FATNetFA}  &  94.41 &  \textbf{97.41} &  97.03 &  89.62 &  94.40 &    30.0 &    23.0 \\
    \midrule
    &   \textbf{Ours} &  96.05 &  96.60 &  \textbf{97.71} &  \textbf{92.30} &  \textbf{95.70} &     \textbf{3.0} &     \textbf{1.3} \\
    \bottomrule
    \end{tabular}
    \label{tab: table4}
    \end{adjustbox}
    \end{subfigure}   
    \bigskip
    \begin{subfigure}{10cm}
        \centering
        \includegraphics[width=8cm, height=4cm]{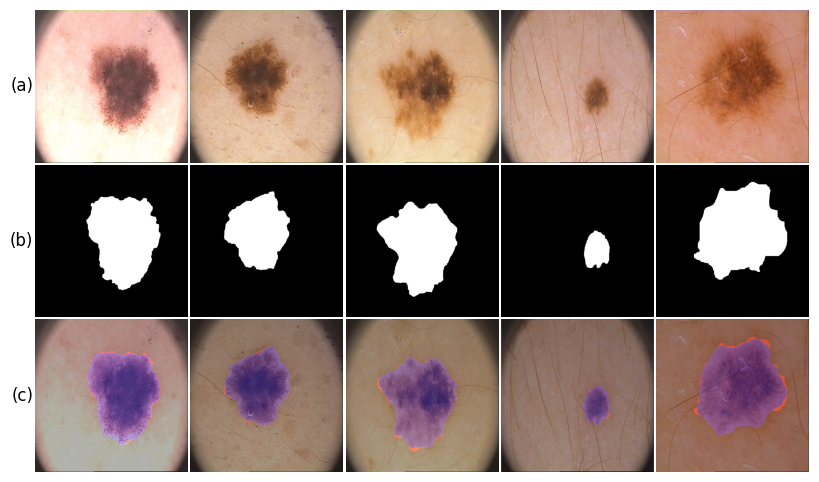}
    \captionsetup{labelformat=empty} 
    \caption{\textbf{Fig. 7:}Qualitative results on PH2 Dataset. (a) Original dermoscopic input (b) Ground truth mask (c) Overlaid predicted mask (blue) and ground truth mask (red) on original image. Qualitative results complements the quantitative findings in terms of segmentation performance.Appearance of red in the overlaid image indicates segmentation differences between the predicted and ground truth.}
    \label{fig: table4fig}
    \end{subfigure}
    \label{my label}
\end{figure*}

The ISIC 2016 dataset represents one of the first standardized skin lesion segmentation task comprising of 900 training images and 300 testing images. Our proposed MobileUNETR is bench marked against nine different architectures, encompassing FCNN, Attention-augmented FCNNs, Generative Adversarial Network (GAN)-based  methods, and Transformer-based methods. Performance results across seven metrics are consolidated in Table 1 with a 2.17\% and 1.21\% increate in IoU and Dice metrics respectively. 

\subsection{Results on ISIC 2017}

The ISIC 2017 improves the scope of skin lesion segmentation by expanding the data corpus size. This dataset comprises 2500 training images with 600 testing images. Our proposed MobileUNETR is systematically bench marked against 12 diverse architectures. We showcase that our model consistently demonstrates improvements in IOU, Dice, and accuracy metrics in all architectures while maintaining a lightweight and efficient design. Results are presented in Table 2 where we boast a 2.47\% and 1.84\% increase in IoU and Dice metrics respectively.

\subsection{Results on ISIC 2018}

The ISIC 2018 dataset stands out as the most comprehensive among commonly utilized skin lesion segmentation datasets. The dataset comprises of 2694 training images together with 1000 testing images. Similar to previous experiments, proposed MobileUNETR is bench marked against a diverse set of 10 architectures, covering a wide range of architectures. Performance outcomes across seven metrics for ISIC 2018 are consolidated in Table 3. Our results consistently reveal enhancements ranging from 2.54\%to 1.71\% in IOU, and Dice metrics across all architectures, while maintaining a lightweight and efficient design.

\subsection{Results on ISIC PH2}

\begin{figure*}[ht]
\centering
    \begin{subfigure}{10cm}
        \centering
    \captionsetup{labelformat=empty} 
        \caption{\textbf{Table 5:} Performance comparison between MobileUNETR and advanced architectures and training methods.}
    \begin{adjustbox}{width=10cm, center}
   \setlength{\tabcolsep}{5pt} 
  \begin{tabular}{lccrrrrrr}
    \toprule
    Model & Params (M) ↓ & GFLOPs ↓ & \multicolumn{2}{c}{IoU ↑} & \multicolumn{2}{c}{Dice ↑} \\
    \cmidrule(lr){4-5} \cmidrule(lr){6-7}
    & & & ISIC & PH2 & ISIC & PH2 \\
    \midrule
    ViT-B w/ PEFT \cite{Du2023AViTAV} & 91.8 & 18 & \textbf{83.71} & 91.72 & 90.77 & 95.64 \\
    AViT w/ PEFT\cite{Du2023AViTAV} & 99.4 & 20.9 & 85.22 & 91.72 & \textbf{91.74} & 95.66 \\
    VPT w/ PEFT \cite{Jia2022VisualPT} & 92.8 & 26.5 & 83.83 & 87.27 & 90.89 & 93.14 \\
    AdaptFormer w/ PEFT \cite{Chen2022AdaptFormerAV} & 93.0 & 18.2 & 84.15 & 88.33 & 91.12 & 93.76 \\
    H2Former\cite{He2023H2FormerAE} & 33.7 & 24.7 & 84.35 & 91.77 & 91.17 & 95.65 \\
    BAT\cite{Wang2021BoundaryAwareTF} & 46.2 & 10.3 & 84.40 & 92.04 & 91.33 & \textbf{95.84} \\
    TransFuse\cite{Zhang2021TransFuseFT} & 143.5 & 64.3 & 85.22 & \textbf{92.69} & 91.73 & 96.18 \\
    \midrule
    Ours & \textbf{3.0} & \textbf{1.3} & 84.59 & 92.25 & 90.65 & 95.76 \\
    \bottomrule
  \end{tabular}
    \label{tab: table5}
    \end{adjustbox}
    \end{subfigure}
    \label{fig: table5fig}
\end{figure*}

Finally, we present the evaluation of MobileUNETR's performance using the PH2 dataset. Unlike the earlier ISIC datasets, PH2 represents a relatively compact dataset, providing an opportunity to highlight the generalization capabilities of our hybrid architecture in handling smaller datasets. Aligning with our previous experiments, we benchmarked the proposed architecture against nine diverse architectures, and performance results are presented in Table 4. Our results consistently reveal improvements ranging from 2.68\% and 1.3\% in IOU and Dice metrics, respectively. Successful experiments on PH2 demonstrate the adaptability of our proposed model for applications involving sparse datasets.

\subsection{Comparison to Advanced Training Techniques}

As an alternative to designing lightweight deep learning architectures a class of advanced training techniques called Parameter Efficient Fine Tuning (PEFT) \cite{Hu2021LoRALA,Dettmers2023QLoRAEF} has been prevalent in recent research. To demonstrates that despite the compact size of the architecture, our model achieves results that rival those of larger architectures employing advanced training techniques we compare our method with recent solutions employing these advanced techniques. Table 5 showcases the effectiveness of well-designed lightweight architectures, proving they can be as effective as large complex models and emphasizing that over-parameterization is not the future of modern deep learning.

\section{Conclusion}

Encoder-decoder architectures provide researchers with a strong architectural paradigm for medical image segmentation. Although it has been used successfully to push the boundaries of medical image segmentation, larger and more complex versions of the encoder decoder paradigm may not be the solution for modern deep learning architectures. This paper introduces MobileUNETR, an innovative and efficient hierarchical hybrid Transformer architecture with tailored for image segmentation. Unlike existing methods, MobileUNETR efficiently integrates local and global information in both the encoder and decoder stages, leveraging the benefits of convolutions and transformers. This integration allows the encoder to extract local and global features during the encoding stage, while allowing the decoder to reconstruct these features, ensuring both local and global alignment in the final segmentation mask. By incorporating local and global features at each level, MobileUNETR avoids the need for large, complex, and over-parameterized models. This not only enhances performance, but also significantly reduces model size and complexity. Extensive experiments were carried out that compared and contrasted our proposed medical image segmentation method with four widely used public datasets (ISIC 2016, ISIC 2017, ISIC 2018, and PH2 dataset). Comparative analyses with state-of-the-art methods demonstrate the effectiveness of our MobileUNETR architecture, showcasing superior accuracy performance and excellent efficiency in model training and inference. Across all datasets MobileUNETR demonstrates a 1.3\% to 2.68\% increase in Dice and IoU metrics with a 10x and 23x reduction in parameters and computational complexity, compared to current SOTA models. We hope that our method can serve as a strong foundation for medical imaging research, since the application of MobileUNETR in image segmentation is endless. Additionally, we hope that our work here has opened the door to motivate further research in efficient architectures in medical imaging research.

\bibliographystyle{splncs04}
\bibliography{references}
\end{document}